# Motor Imagery EEG Signal Classification Using Minimally Random Convolutional Kernel Transform and Hybrid Deep Learning

Jamal Hwaidi[a], Mohamed Chahine Ghanem[b,*]

[a]Department of Electrical and Electronic Engineering, City University of London, EC1V 0HB, London, UK
[b]Department of Computer Science, University of Liverpool, L69 3BX, Liverpool, UK

ABSTRACT

The brain-computer interface (BCI) establishes a non-muscle channel that enables direct communication between the human body and an external device. Electroencephalography (EEG) is a popular non-invasive technique for recording brain signals. It is critical to process and comprehend the hidden patterns linked to a specific cognitive or motor task, for instance, measured through the motor imagery brain-computer interface (MI-BCI). A significant challenge is presented by classifying motor imagery-based electroencephalogram (MI-EEG) tasks, given that EEG signals exhibit nonstationarity, time-variance, and individual diversity. Obtaining good classification accuracy is also very difficult due to the growing number of classes and the natural variability among individuals. To overcome these issues, this paper proposes a novel method for classifying EEG motor imagery signals that extracts features efficiently with Minimally Random Convolutional Kernel Transform (MiniRocket), a linear classifier then uses the extracted features for activity recognition. Furthermore, a novel deep learning model based on Convolutional Neural Network (CNN) and Long Short-Term Memory (LSTM) architecture was proposed and demonstrated to serve as a baseline. The classification via MiniRocket's features achieved higher performance than the best deep learning models at a lower computational cost. The PhysioNet dataset was used to evaluate the performance of the proposed approaches. The proposed models achieved mean accuracy values of 98.63% and 98.06%, respectively, for the MiniRocket and CNN-LSTM. The findings demonstrate that the proposed approach can significantly enhance motor imagery EEG accuracy and provide new insights into the feature extraction and classification of MI-EEG.

## 1. Introduction

A human-computer interaction technique based on brain signals is known as brain-computer interface (BCI) technology [1–3]. It offers a communication channel for non-neuromuscular control and communication between the human brain and the outside world using a brain-computer interface without the use of muscles or the peripheral nervous system [4].

Electroencephalography (EEG) signals represent electrical signals from the brain nerves in the BCI system. It serves as the system's foundation for signal processing as well. The electrical signals produced by the brain's neurons during EEG brain rhythms are microvolts [5]. EEG uses affordable equipment and permits patient movement while recording. These are advantages over other non-invasive recording methods such as magnetoencephalography (MEG) and functional magnetic resonance imaging (fMRI), which require patients to remain stationary while using expensive, large-scale equipment [6].

Various EEG signal types have been employed as BCI control signals. The most common signals are P300 evoked potentials, steady-state visual evoked potentials (SSVEP), and motor imagery (MI). The power spectrum of various frequency bands can change for various movement tasks, reflecting neuronal firing pattern changes. Event-related synchronisation (ERS) and event-related desynchronisation (ERD) are two names for this phenomenon. The primary spectrums of ERS and ERD in MI tasks are $\mu$ (8-14 Hz) and $\beta$ (14-30 Hz) [7].

Different kinds of motor or cognitive activities can be understood using EEG. The term "motor imagery" (MI) describes a subject's ability to move their limbs mentally even though they are not being moved [8].

An emerging area of biomedical applications is BCI based on EEG motor imagery [9]. The motor imagery classification problem can also be used to identify and separate various types of thinking or imagining [10, 11]. Due to its advantages over other cerebral signals, researchers have given special consideration to the classification of EEG signals. Although a specialist could also perform this classification, doing so would not be as quick or accurate, so various machine learning and deep learning techniques have been used to automate the classification of MI.

Traditional MI-BCI classification approaches are generally categorised into two groups based on the features of EEG signals: (1) spatial feature classification [12] and (2) spatial-frequency feature classification [13]. According to Sakhavi et al. [14], several prior MI-BCI algorithms neglected the temporal characteristics of EEG signals and neglected to consider the dynamic energy representation of EEG. Automatic function selectors are only partially effective at reversing this tendency.

Deep learning (DL) techniques have recently outperformed traditional handcrafted techniques in a number of fields, including image processing, speech processing, video processing, and text processing [15–17]. EEG signal characteristics like low signal-to-noise ratio (SNR), fewer data, and multiple channels make it challenging to develop a general DL model for the identification of EEG signals.

* Dr M. C. Ghanem is the corresponding author.
jamal.hwidi@city.ac.uk (J. Hwaidi);
mohamed.chahine.ghanem@liverpool.ac.uk (M.C. Ghanem)
ORCID(s):

It is important to have a meaningful design for the DL framework used to process EEG signals. Some successes have been seen in researchers' efforts to apply DL techniques to the BCI field.

Deep neural networks (DNN) with more data, improved learning methods, and faster computation have gained in popularity [18]. By processing the data from an EEG time series in the RNN networks' internal memory, EEG signals can be decoded [19]. As opposed to a normal RNN, Long Short-Term Memory (LSTM) is an RNN with the capacity to maintain the sequence of data over an extended period and identify the desired pattern [20]. Convolutional neural network (CNN) models can recognise strong spatial details in images [21]. Researchers have extensively applied to investigate the classification and spatial characteristics of EEG signals [22–24]. To improve DNNs' capacity to simultaneously extract spatial and temporal characteristics, CNN and LSTM neural networks were combined to create a hybrid neural network that can learn both spatial and temporal features [25].

Despite the impressive accomplishments of prior work in this field, the BCI system still lacks standards for practical application. Moreover, there is still considerable opportunity for improvement in both the EEG signals classification methodology and accuracy.

The main contributions of this paper are as follows:

(1) The independent component analysis (ICA) technique is utilised to separate the $\mu$ and $\beta$ frequencies as segmentation targets from other frequencies.
(2) The MI EEG signals are subjected to a Minirocket feature extraction method. The MiniRocket can enhance the accuracy of EEG classification by retaining more crucial feature information while also processing less data.
(3) A hybrid neural network, which consists of CNN and LST,M is utilised to enhance motor imagery accuracy in EEG classification and compared to MiniRocket's feature-based classifier.

The remainder of the paper is organised as follows: Section 2 reviews the literature on the classification of EEG MI utilising machine learning and deep learning techniques. Section 3 presents the data description and system method. The discussion of the methodology with existing work and conclusion are given in Sections 4 and 5, respectively.

## 2. Related Work

This section examines a variety of traditional machine learning and deep learning techniques used in the classification of motor imagery for the BCI. Substantial efforts have been made in the past to improve the accuracy of the MI classification through feature extraction and classification algorithms [26, 27].

To classify MI-EEG data, conventional machine learning techniques have been extensively used. The MI-EEG signal is typically processed using traditional methods in three steps: preprocessing, feature extraction, and classification.

Preprocessing involves a number of operations, as channel selection, signal filtering, signal normalisation, and removal of artefacts (removing the noise from MI-EEG signals). Independent component analysis (ICA) is the technique that is most frequently used to remove artefacts [28].

Existing research indicates that a variety of feature extraction algorithms have been proposed for extracting task-related MI features from EEG signals with high dimensions. Depending on the processing domain for the data, the MI features can be categorised into three types: temporal features, spectral features, and spatial features. Temporal features like mean, variance, Hjorth parameters, and skewness are derived from the time domain at various time points or throughout various time segments [29]. Spectral features can be either time-frequency features like short-time Fourier transform (STFT) [30] and wavelet transform (WT) [31, 32] or frequency-domain features like power spectral density (PSD) and fast Fourier transform (FFT) [33]. Spatial features, such as common spatial patterns (CSP), are intended to identify characteristics from specific scalp electrode locations [34, 35]. The regularisation feature in sparse CSP [36] adds sparsity to CSP values. Other methods that have been tried to improve CSP functionality include stationary CSP [37], divergence CSP [38], sub-band CSP (SBCSP) [39], probabilistic CSP [40], and frequency domain CSP (FD-CSP) [41]. Filter bank CSP (FBCSP) [42] is an additional enhanced variation of the CSP approach that makes use of the frequency data in MI-EEG signals as well as the spatial information in EEG channels. Furthermore, discriminative filter bank CSP (DFBCSP) [43] is an extension of CSP that uses optimisation to create spatial weights for finite impulse response filters.

During the classification phase, various classifiers were used, for instance, the naive Bayesian classifier [44], linear discriminant analysis (LDA) [45], support vector machine (SVM) [46], K- nearest-neighbour (KNN) [47], and extreme learning machine (ELM) [48, 49], to categorise the derived MI features into different MI tasks.

Sharma et al. [50] used a neighbourhood component analysis (NCA)-based supervised learning framework to enhance the spatial features that CSP had obtained for each sub-band. SVM was ultimately trained to handle the classification. SVM is inefficient when there are a large number of observation samples and is sensitive to missing data, despite the fact that it can perform inference in a reasonably quick amount of time. KNN's use in classifying MI EEG signals is constrained by its high time and space complexity, which is not sensitive to missing values. Generally, even though the traditional machine learning algorithms have produced impressive results, their accuracy has not been satisfactory.

On the other hand, being able to automatically extract the discriminative features, deep learning methods with deep neural networks (DNN) have recently performed remarkably well in several areas, such as speech recognition, computer vision, and medical diagnosis [51–56]. The growing popularity of deep learning has significantly reduced the need for feature extraction. Deep learning algorithms can produce and predict new features without specifying which ones to use or how to extract them [57, 58].

To classify MI EEG, various researchers have used deep learning methods. CNN can learn reliable spatial features [59]. RNN can successfully learn the EEG signal's sequential relationship [60]. For unsupervised feature learning, autoencoder (AE) models are appropriate [61]. Shallow CNN (sCNN) and deep CNN (dCNN) architectures were developed by Schirrmeister et al. [59] to extract a wide range of features and to comprehend the temporal structure of band power variations. However, the dCNN underperformed the sCNN because there was less training data available. Studies have indicated that the CNN's various layers can extract various abstractions from the EEG signal. As a result, Amin et al. [22] layered and filtered extracted features from CNN models to discover how time-series signals are structured both locally and overall. The classification accuracy was greatly improved by combining features. Moreover, the pragmatic CNN (pCNN) network was developed by Tayeb et al. [62], which is less complicated and more accurate than dCNN. Tabar and Halici [63] developed a network to classify two classes of MI tasks using CNN and stacked autoencoders (SAE). They first used Short-Time Fourier Transform (STFT) to extract information from the EEG signal's time and frequency domains and convert it into two-dimensional (2D) images, forwarding these images to the CNN-SAE network for classification. To improve differential phase-synchronised representations for examining the EEG-synchronised transitions from being alert to being drowsy, Reddy et al. [64] used fuzzy common-space patterns. The findings of the data analysis indicate that traditional regression methods like support vector regression (SVR) and ridge regression are outperformed by the suggested multi-task deep network. Four expert CNNs were combined into a modular network by Brenda et al. [65]. Binary classification is carried out by each expert CNN, and the results are then fully connectedly analysed.

EEG signals are time-sequenced nonlinear signals. Time series with nonlinear characteristics, like EEG, are frequently handled by long-short term memory (LSTM), a RNN variant. The LSTM network was employed for the first time by Shen et al. [66] to deal with issues with EEG signal classification, and they achieved notable improvements in both the MI-EEG dataset used in BCI competitions with individuals who are healthy, as well as the dataset compiled from stroke patients. Speech recognition and natural language processing were the inspirations for this work. The LSTM deep neural network was employed by Tortora et al. [67] in order to deal with time-dependent information contained in brain signals during locomotion. Lu et al. [68] classified sequential-spatial-frequency features using a deep RNN architecture. Additionally, by viewing the signal as a various non-stationary and nonlinear characteristics, this method also reduces variations in classification between individuals. Ma et al. [69] extracted both spatial and temporal properties from the *eegmmidb* dataset by combining the LSTM and bidirectional LSTM (BiLSTM), obtaining a result that was 8.25% more accurate than conventional methods.

Extracting both spatial and temporal information simultaneously, a variety of neural networks have been combined (hybrid DNN). Zhang et al. [70] employed CNN and LSTM to extract separately the signal's sequence relationship and frequency-spatial information. Due to their ability to simultaneously learn many different features, hybrid neural networks were shown to perform better than other neural networks. Li et al. [71] proposed a CNN- and LSTM-based multilevel spatial-temporal feature fusion EEG classification algorithm. Integral extraction and fusion of the middle layer feature, the spatial feature, and the temporal feature. It overcomes the traditional machine learning algorithms' flaw of easily losing manually chosen features. Furthermore, Khademi et al. [72, 73] proposed a hybrid neural network using CNN and LSTM to classify four motor imagery tasks in the BCI Competition IV dataset 2a.

## 3. Methodology

Fig 1 illustrates the proposed model architecture. A total of 10 datasets were downloaded, five for the training set, two for the validation set, and the remaining three for the test set. These data were downloaded separately for each subject, resulting in 10 data points in total. Data from various subjects were combined to create the test and training set. Five symmetrical electrode pairs placed throughout the motor cortex produced the MI-EEG raw signals for each trial, with the signals from each pair constituting a sample.

The MiniRocket Features (MRF) module is then used to send the denoised signal to the Minirocket algorithm. MRF extracts the features from a minibatch of time series samples. A linear classifier can model the information in the extracted features related to the series class membership. Moreover, A hybrid CNN-LSTM architecture was also proposed. As CNN-LSTM uses two convolutional layers for feature extraction and an LSTM layer for additional classification, these architectures offer a comparable baseline. This allowed us to compare the effectiveness of randomised kernels (MRF) and deep learning-based methods (CNN-LSTM).

### 3.1. Dataset

The BCI2000 system developers [74] recorded the PhysioNet MI-EEG dataset [75]. It is made up of more than 1,500 EEG recordings lasting between one and two minutes that were recorded at a sampling rate of 160 Hz from 109 various subjects. Each subject performed four MI tasks: T1, T2, T3, and T4, referring to the left and right fists, both fists, and both feet, respectively. Each MI task required 21 trials. Fig 2 shows the timing diagram for the trial. The subject relaxes for 2 seconds before the start of the trial at time $t = -2$. At time $t = 0$, the target appears on the screen, and the symbols are as follows:

(1) L stands for the left fist motor imagination as it opens and closes.
(2) R stands for the right fist motor imagination when it opens and closes.
(3) BLR stands for both fists' motor imagination when it opens and closes.
(4) BF stands for both feet motor imagination when it opens and closes.

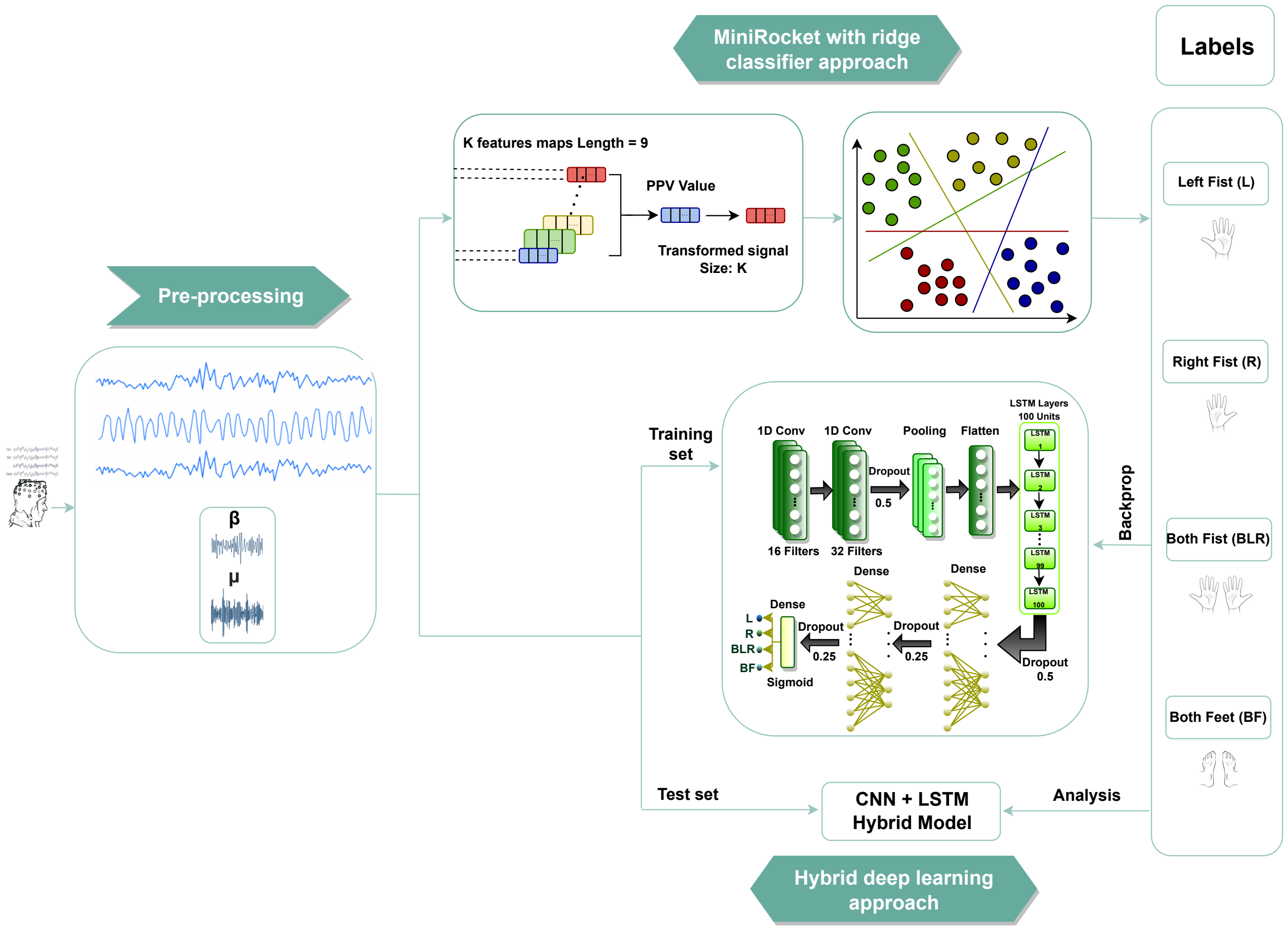

**Figure 1:** The proposed model architecture.

The subject received the MI task for four seconds. The trial ends when the target disappears at $t = 4$ seconds. A new trial starts after a two-second intermission [76]. The motor imagination operates at a sampling frequency of 160 Hz for approximately four seconds each time. Therefore, the effective data size for each electrode during a test is 640. The sample contains a pair of symmetric electrodes, and the data from these electrodes is serially connected. The sample size is therefore 1,280.

For each MI task, each subject completed 21 trials for a total of 84 trials. A 10-fold cross-validation was performed on the dataset used in this study. The trials consisted of 10 segments for each subject. Two test trials and the remaining training trials were used for each task class, each of which had its own requirements. Thus, the training set has 76 trials, while the test set has 8 trials. There are 840 trials total, 760 for training and 80 for testing, in the datasets for 10 subjects (S1~S10). Additionally, 9 samples were produced by each trial. In this study, a dataset containing 7,560 samples from 10 subjects was chosen for model training and generalisation performance validation.

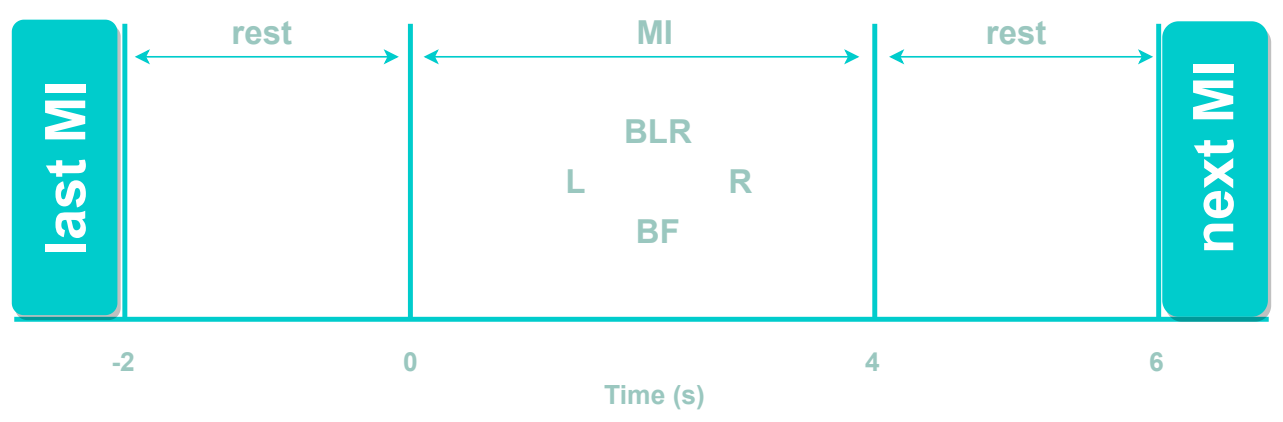

**Figure 2:** The trial's timing diagram.

## 3.2. Preprocessing

Signal amplification and filtration processes are applied to the data at the time of acquisition. The use of data segmentation was demonstrated to segment the data stream. The EEG dataset was preprocessed using the following procedures:

1. A 128 Hz sampling rate was used for the data.
2. In this experiment, $\mu$ and $\beta$ waves were taken into consideration when dividing EEG signals into bands using independent component analysis (ICA).
3. 60-second trials were used to divide the data, so three-second pre-trial intervals were eliminated, and the data was averaged using a common reference.

## 3.3. MiniRocket

High computational complexity is a limitation of the majority of state-of-the-art (SOTA) time series classification techniques. They are consequently difficult to train on smaller datasets and virtually useless on larger datasets. In comparison to other SOTA time series classifiers, Random Convolutional Kernel Transform (ROCKET) [77] recently achieved SOTA accuracy in a much shorter amount of time by modifying the deep learning convolutional kernel to enhance the precision of shapelet extraction algorithms, which are used to extract readable characteristic units of variation from time series [78].

Rocket applies 10,000 random convolutional kernels to the input time series transformation (random in terms of their length, weights, bias, dilation, and padding). Then, it computes two features from each convolution output using the proportion of positive values (PPV) and max pooling operators, yielding 20,000 features per time series. A linear classifier is trained using the transformed features. The two main components of Rocket that help it achieve SOTA accuracy are the use of dilation and PPV.

Minimum Random Convolutional Kernel Transform (MiniRocket) [79] further lessens the algorithm's computational complexity by restricting the variance to only positive variance, and only those parameters that can be characterised by a variety of time series data are used in the convolutional kernel. Rocket's recommended default variant is MiniRocket because it is faster and has Rocket's level of accuracy. A representation of the ROCKET architecture for extracting features from an input signal is shown in Fig 3.

Rocket and MiniRocket both use PPV of the convolution results to pool convolution values, which is the most important component and a factor in their high accuracy. PPV is defined as:

$$PPV = \frac{1}{m}\sum_{i=1}^{m}\left[x_i \odot c_i + b > 0\right] \tag{1}$$

where $c_i$ is the random convolutional kernel applied to the $i$th time sequence, $x_i$ is the $i$th time sequence, $\odot$ is the inversion bracket, and $b$ is the bias scalar.

Additionally, unlike the maximum value of a feature map calculated by the ROCKET method, which is used as a classification feature, the PPV is the only feature extracted by MiniRocket. Because of this, the size of the transformed signal is reduced to $K$, whereas in the ROCKET version it was doubled. The default number of generated kernels is 10,000, but this can be adjusted manually if necessary. The authors made some additional optimisations, which almost eliminated the randomness in the computation time. In the default configuration, 32 dilations are allowed per kernel. On the other hand, we found that the best results came from using a maximum of 28 dilations per kernel in our experiments.

The computational complexity for an input size of length $I$, the total number of generated kernels $K$, and total samples $S$ is given by $O(I.K.S)$. This is because, unlike the weights that are learned in convolutional-based neural networks, the weights in this method do not change. Instead, many kernels that each have their own set of attributes ensure that a wide variety of features that are related to the input signal are considered.

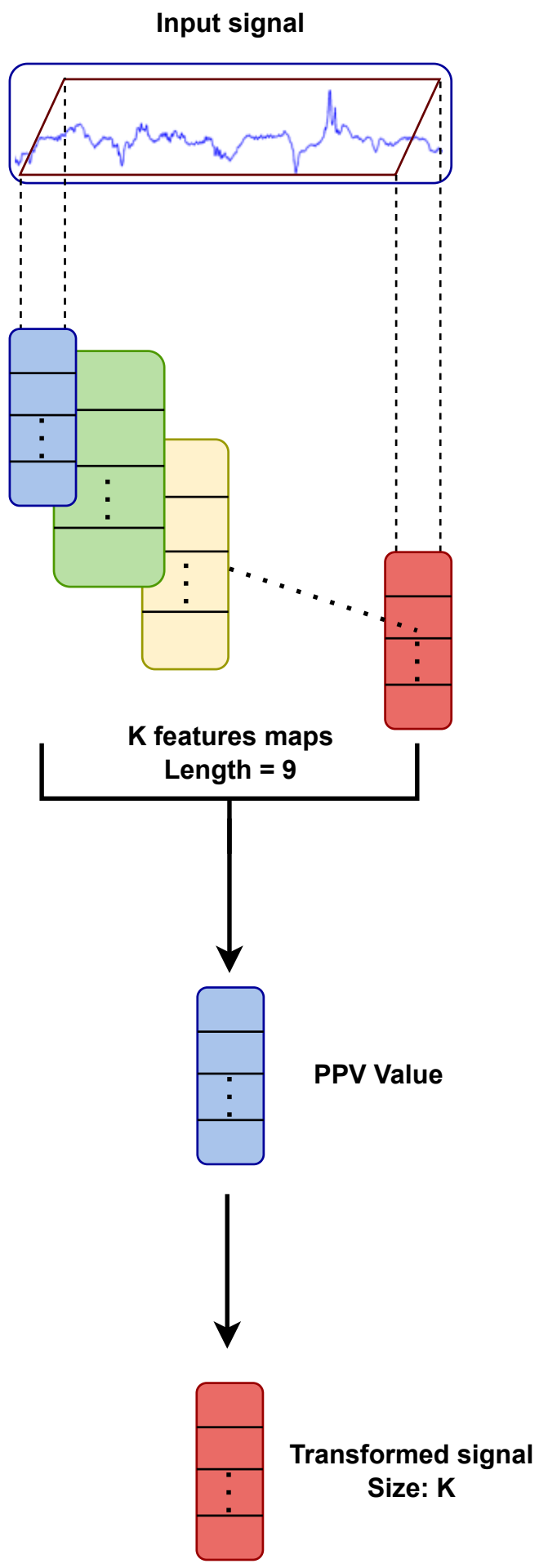

**Figure 3:** MiniRocket framework.

### 3.4. Convolutional Neural Network (CNN)

CNN has emerged as a widely used deep learning-based network for identifying features across various tasks. In contrast to conventional machine learning algorithms, the CNN does not require manual feature design. Instead of losing valuable information, it simply utilises the convolution kernel's local receptive field to learn abstract features from the original data for classification. In comparison to the conventional framework, where feature learning and classification are typically done in two separate steps, the CNN uses multilayer neural networks to simultaneously learn features and classify.

The five layers of a standard CNN are as follows: input layer, convolutional layer, activation layer, pooling layer, and fully connected layer. A classifier layer is usually placed after one or more of these blocks to create a complete CNN [80]. Consider a CNN with $n$ computational blocks and a 2-D or 3-D array as input data $\boldsymbol{J}$. The output of the convolutional layer in the $L$th block can be generated by

$$f_L(\boldsymbol{J}) = \sum_{i=1}^{L} \left( \boldsymbol{J}^i \odot w_i + b_i \right) \tag{2}$$

where $\boldsymbol{J}^i$, $w_i$ and $b_i$ stand for input, weights, and bias, respectively, and $\odot$ denotes the convolution operation. The output of a convolutional layer is taken into consideration as an input of the subsequent block after computational operations on the corresponding activation and pooling layers.

The final convolutional layer's output is fed sequentially into the fully connected and classifier layers. A supervised loss function is back-propagated to train the entire network.

### 3.5. Long Short Term Memory (LSTM)

LSTM is frequently employed to address the characteristics of time series that are nonlinear. Long and short-term dependencies in sequential data can be learned using a memory cell $c$, which has self-connections to store the network's temporal state [81]. Fig 4 shows that each LSTM unit receives three inputs for information processing: $x_t$ serves as the time step's input at this moment, The output of the preceding LSTM unit is $h_{t-1}$, and $c_{t-1}$ is the preceding unit's cell state.

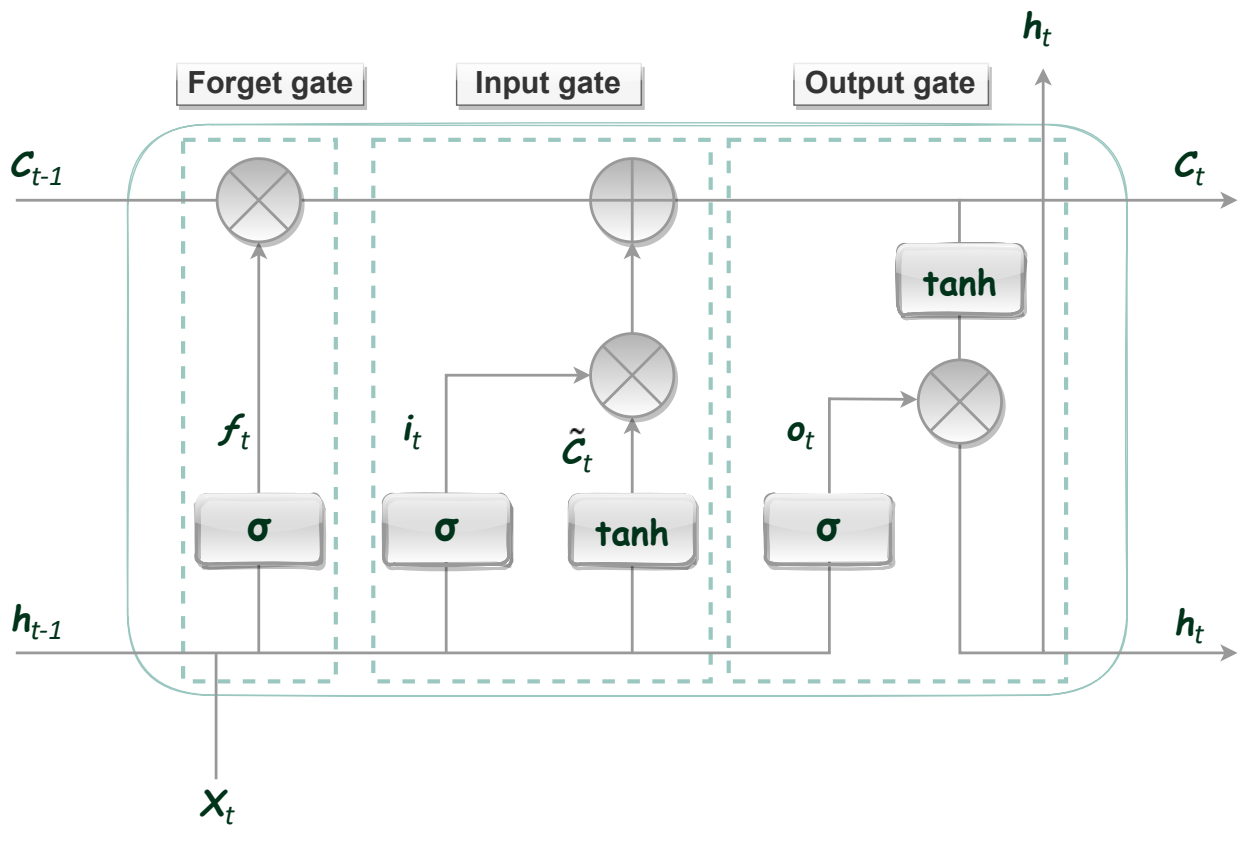

**Figure 4:** The structure of an LSTM unit.

The LSTM unit is controlled by three gates, namely the forget gate, the memory cell, and the output gate, which significantly enhance LSTM's capacity to process temporal data. The data that would be removed from the previous cell is determined by the forget gate and is provided as

$$f_t = \sigma \left( W_f \left[ h_{t-1}, x_t \right] + b_f \right) \tag{3}$$

The memory cell takes the short-term memory $\tilde{c}_t$ and the long-term memory, it outputs them from the forget gate, merging them immediately.

$$i_t = \sigma \left( W_i \left[ h_{t-1}, x_t \right] + b_i \right) \tag{4}$$

$$\tilde{c}_t = \tanh \left( W_c \left[ h_{t-1}, x_t \right] + b_c \right) \tag{5}$$

$$c_t = f_t \odot c_{t-1} + i_t \odot \tilde{c}_t \tag{6}$$

The output gate is the final method for obtaining the output $h_t$

$$o_t = \tanh \left( W_o \left[ h_{t-1}, x_t \right] + b_o \right) \tag{7}$$

$$h_t = o_t \odot \tanh \left( c_t \right) \tag{8}$$

where $x_t$ is the hidden layer's input and $x_{t-1}$ stands for the previous cell's output. The bias value is shown by $b$, while $w$ displays the weight matrix, and the sigmoid function $\sigma$ determines how much of the information will be forgotten.

### 3.6. CNN-LSTM Hybrid model

We used hybrid neural networks made up of CNN and LSTM to simultaneously learn the spatial and temporal properties of MI signals, as shown in Fig 5. Convolutional layers are employed in the first layers of CNN-RNN models to extract features and identify patterns. The RNN layers are then applied with their outputs. In comparison to RNNs, convolutional layers experimentally extract the local and spatial patterns of EEG signals more effectively. Additionally, incorporating convolutional layers into RNN enables a more accurate analysis of the data.

The architecture of the hybrid neural network model is further explained in Table 1. The CNN and LSTM layers make up the 13 layers of this network. The proposed model's first 10 layers consist of two convolutional layers, four dropout layers with varying dropout rates, a max-pooling layer, a flatten layer, an LSTM layer, and two dense layers with ReLU and sigmoid activation functions. The 11th

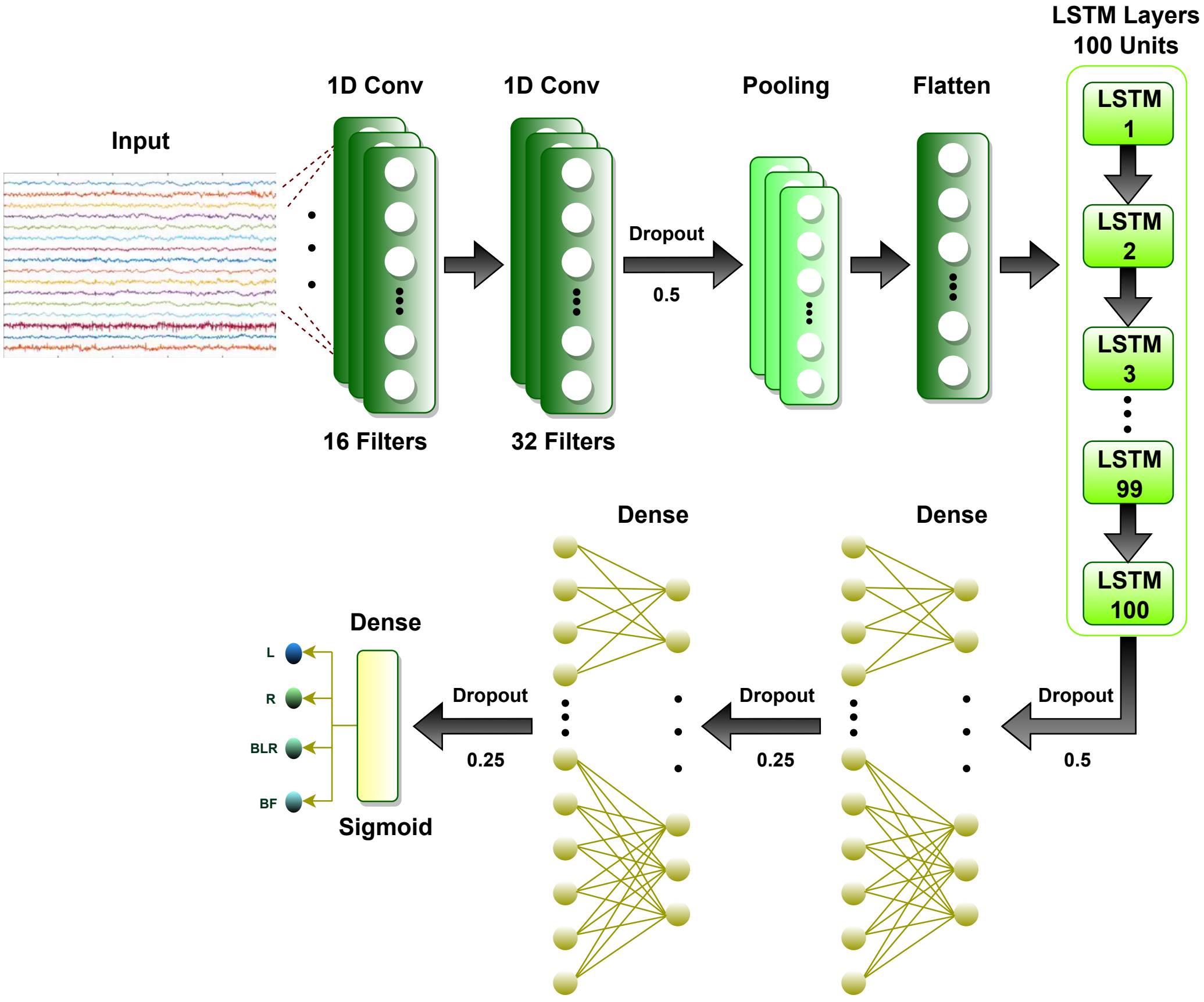

**Figure 5:** The proposed CNN-LSTM hybrid model.

**Table 1**
proposed CNN+LSTM Architecture.

| Layer | Type | Filters | Kernel Size | Stride | Padding | Activation |
|---|---|---|---|---|---|---|
| L1 | Input | - | - | - | - | - |
| L2 | **Conv1D** | 16 | 3 | 1 | VALID | ReLU |
| L3 | **Conv1D** | 32 | 3 | 1 | VALID | ReLU |
| L4 | Dropout | - | - | Rate = 0.5 | - | - |
| L5 | Max Pooling | - | 2 | 1 | VALID | - |
| L6 | Flatten | - | - | - | - | - |
| L7 | **LSTM** | 100 | 62 | - | VALID | - |
| L8 | Dropout | - | - | Rate = 0.5 | - | - |
| L9 | Dense | 100 | - | - | VALID | ReLU |
| L10 | Dropout | - | - | Rate = 0.25 | - | - |
| L11 | Dense | 50 | - | - | VALID | ReLU |
| L12 | Dropout | - | - | Rate = 0.25 | - | - |
| L13 | Dense | 4 | - | - | VALID | Sigmoid |

layer of this architecture uses a dense layer with 50 neurons and the ReLU activation function. Dropout with a rate of 0.25 is present in the 12th layer. Finally, for classification, the 13th layer employs the dense layer with a sigmoid activation function.

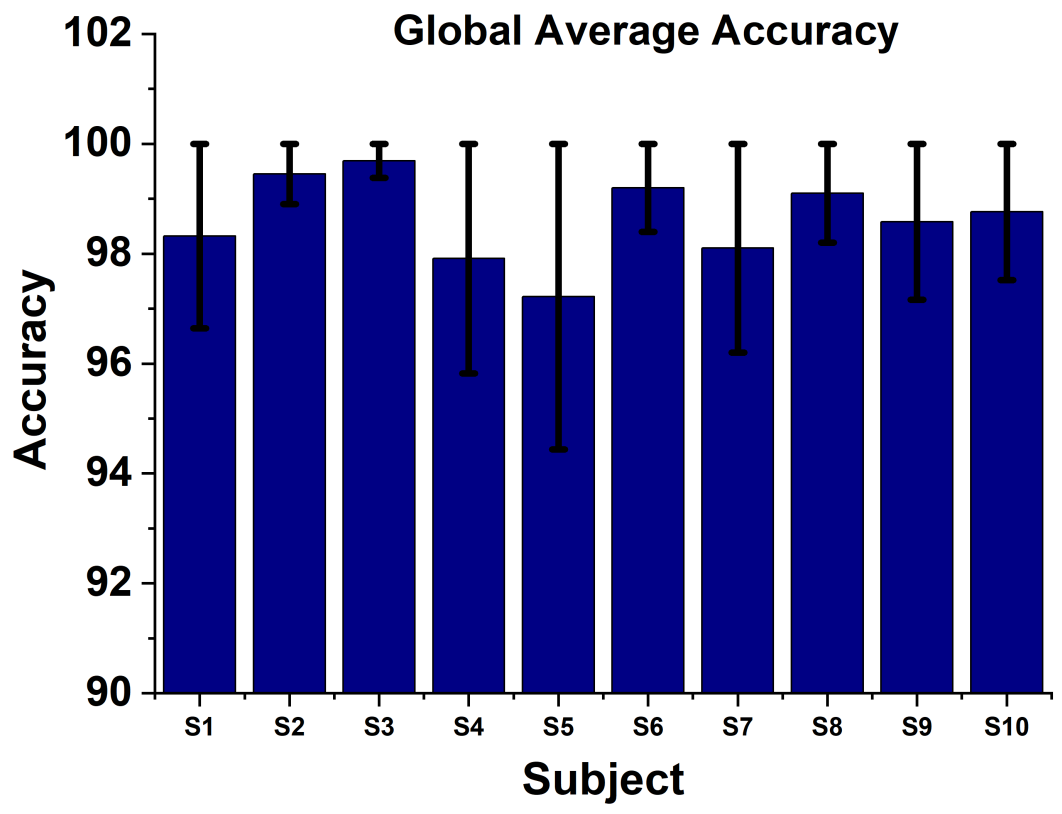

**(a)** The global average accuracy of 10 subjects on the MiniRocket approach

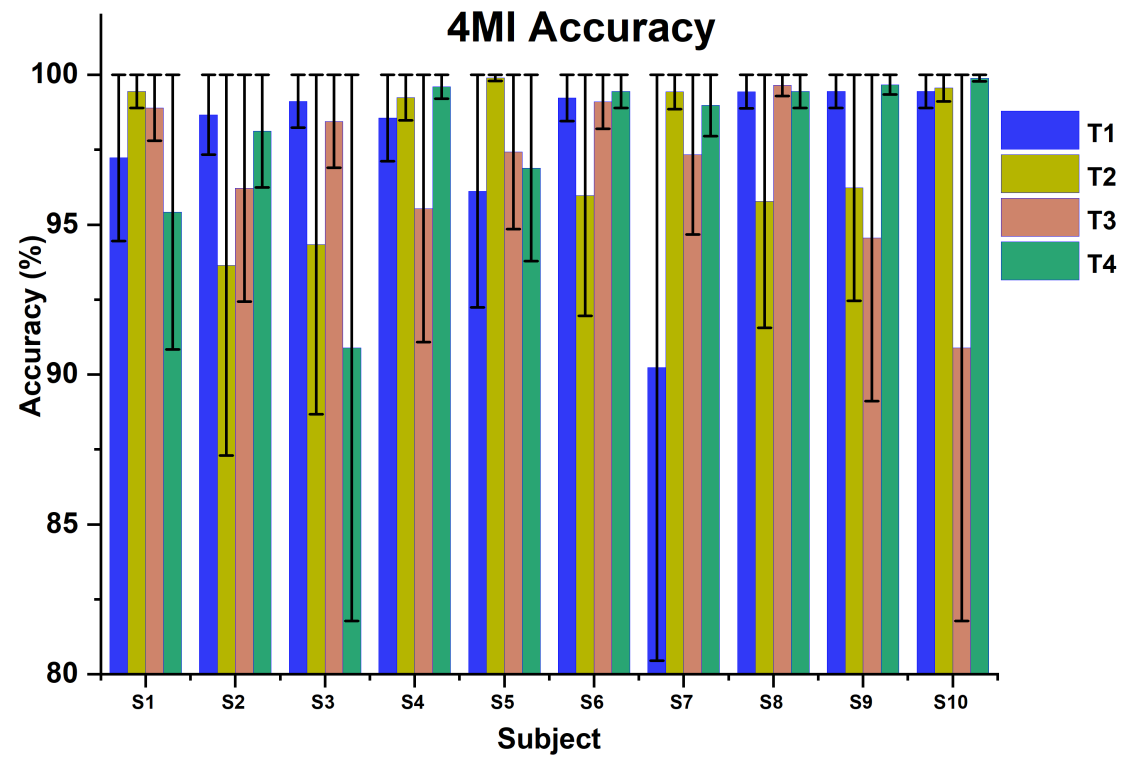

**(b)** The 4 MI accuracy of 10 subjects on the MiniRocket approach

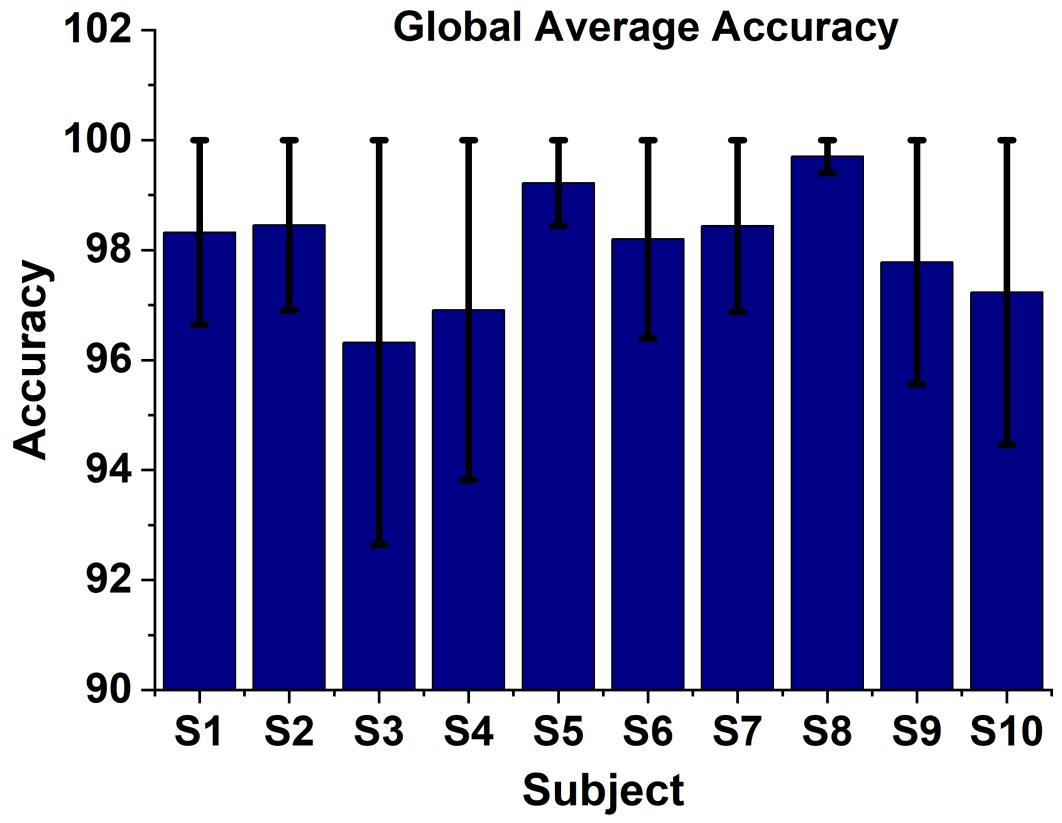

**(c)** The global average accuracy of 10 subjects on the CNN-LSTM approach

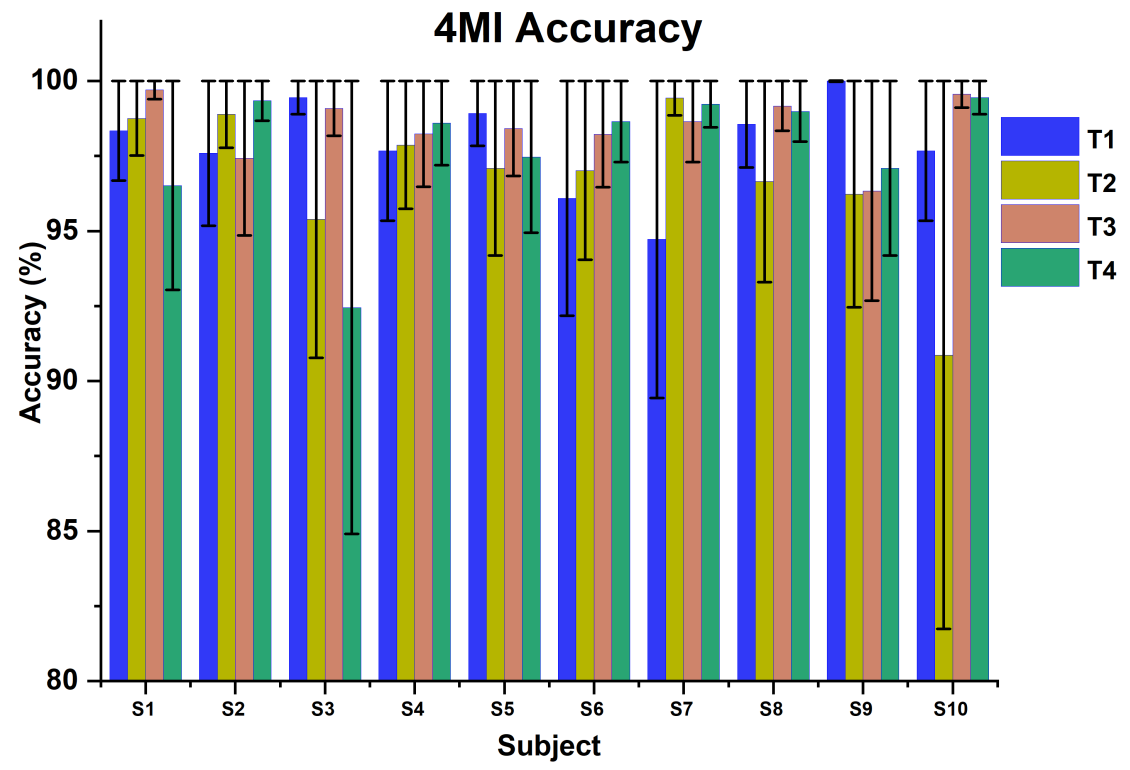

**(d)** The 4 MI accuracy of 10 subjects on the CNN-LSTM approach

**Figure 6:** The global average accuracy and the four tasks of MI accuracy for both approaches on the PhysioNet dataset.

## 4. Experimental Results

The data used was trained and tested on the proposed networks on the Python 3.8 platform, and the experiments in this study were carried out in Python environments on an Intel® Core$^{TM}$ i7-8550U CPU running at 1.80 GHz with processing stages using 16 GB RAM.

In order to evaluate the performance of the proposed model, the classification accuracy ($Ac$) and the ROC curve were used in this paper.

The AUC scale, which runs from 0.5 to 1, represents the area under the ROC curve. Precision, recall, and F-score were used to evaluate the model's performance in identifying four different types of MI. The model performs better for larger values. True positives (TP), true negatives (TN), false positives (FP), and false negatives (FN) are used here.

$$Ac = \frac{TP + TN}{TP + TN + FP + FN} \tag{9}$$

$$Precision = \frac{TP}{TP + FP} \tag{10}$$

$$Recall = \frac{TP}{TP + FN} \tag{11}$$

$$F1 = 2 * \frac{Precision * Recall}{Precision + Recall} \tag{12}$$

The combined data from the two sessions for each subject is split into three sets: a training set, a validation set, and a test set in the ratios of 5:2:3. Adjusting network parameters involves using the training set and the validation set. Therefore, only the test set is utilised to evaluate final performance and is not used for network training.

The training batch size is 64, which represents the volume of data used to update model coefficients during each sub-epoch. The backpropagation (BP) algorithm was used to train the CNN-LSTM hybrid model, which updates each network parameter (weights and biases) iteratively upwards, starting at the bottom up to the top layers to reduce the cost function.

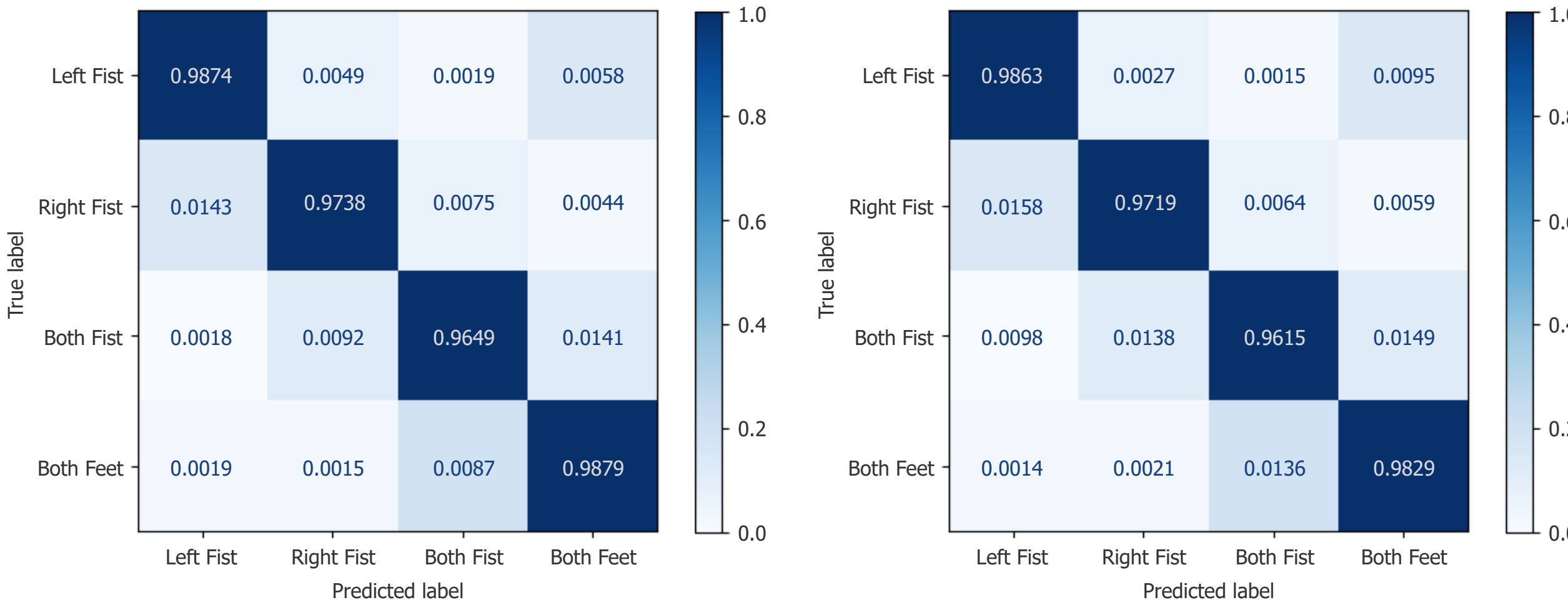

**(a)** The MiniRocket approach mean confusion matrices

**(b)** The hybrid CNN-LSTM approach mean confusion matrices

**Figure 7:** The mean confusion matrices for all subjects from different models.

Rectified linear units (ReLu) were selected as the activation functions because they have a lower probability of gradient vanishing during training and have a faster convergence rate.

$$f(x) = \begin{cases} 0, & \text{if } x > 0 \\ x, & otherwise \end{cases} \tag{13}$$

The adaptive moment estimation (Adam) algorithm accelerates the decay of minimising the cost function globally by adaptively estimating the first-order moment (the mean) and second-order moments (the uncentered variance). In this study, to minimise the loss function, the Adam optimisation algorithm used a persistence of $1 \times 10^{-5}$ as the network's learning rate.

Two regularisation methods, dropout and weight regularisation, were used to prevent overfitting. Dropout was specifically used after each 0.5 dropout LSTM and convolutional layer, as well as after dense layers with a dropout value of 0.25. Using L2 regularisation with a value of 0.01, weight regularisation was applied to all of the architecture's convolutional, LSTM, and dense layers.

The LSTM model consists of 100 LSTM units, and backpropagation through time (BPTT) uses the learning task, which employs $T$ time steps to train the units collectively. The LSTM gated structure allowed it to successfully reduce the vanishing gradient descent. In particular, the forget gate, which enables the network to more effectively control the gradient values at each time step, avoiding convergence to zero, is contained in the gradient, which contains the combined activation vector for all three gates. The training process has been accelerated by using batch normalisation.

To demonstrate the effectiveness of the proposed approaches, an offline study is conducted with the PhysioNet dataset. To ensure that no data blocks were split between the training and test sets, each trial was divided into ten parts: nine for training and one for testing. After that, the model was trained and tested to determine its accuracy. 10 cycles were completed for each subject, starting with data segmentation and ending with training and testing. Their average is used to determine the accuracy across all subjects, on average.

Figure 6(a) illustrates the global average accuracy of 10 subjects in the PhysioNet dataset based on the MiniRocket and linear classifier model, which is 98.63%. The best classification result was S8. Its MI accuracy rates for T1, T2, T3, and T4 are 99.44%, 95.78%, 99.65%, and 99.45%, respectively. The average accuracy of S3 is the lowest, Its MI accuracy rates for T1, T2, T3, and T4 are 99.12%, 94.34%, 98.45%, and 90.89%, respectively, as shown in Fig 6(b). The global average accuracy of 10 subjects in the PhysioNet dataset based on the hybrid deep learning CNN-LSTM model, shown in Fig 6(c), is 98.06%. The classification result for S8 was the best. For T1, T2, T3, and T4, it has MI accuracy rates of 98.56%, 96.65%, 99.17%, and 98.99%, respectively. The average accuracy of S10 is the lowest; according to Fig 6(d), its MI accuracy rates for T1, T2, T3, and T4 are 97.67%, 90.87%, 99.56%, and 99.45%, respectively.

The classification confusion matrices accurately depict the model's performance in classifying various classes. Fig 7(a) and (b) display the results of the group-level classification for the mean confusion matrices for all subjects in both approaches. The different hues of the legend colour stand in for accuracy, while the ratios of classification accuracy and misclassification are shown by the diagonal and non-diagonal lines, respectively. Both the left fist and both feet had the best MI discrimination.

The ROC diagrams of the proposed classification algorithms for the raw input EEG signals from PhysioNet on the MiniRocket and CNN-LSTM approaches are shown in Fig 8(a) and (b). The model's performance at classifying data

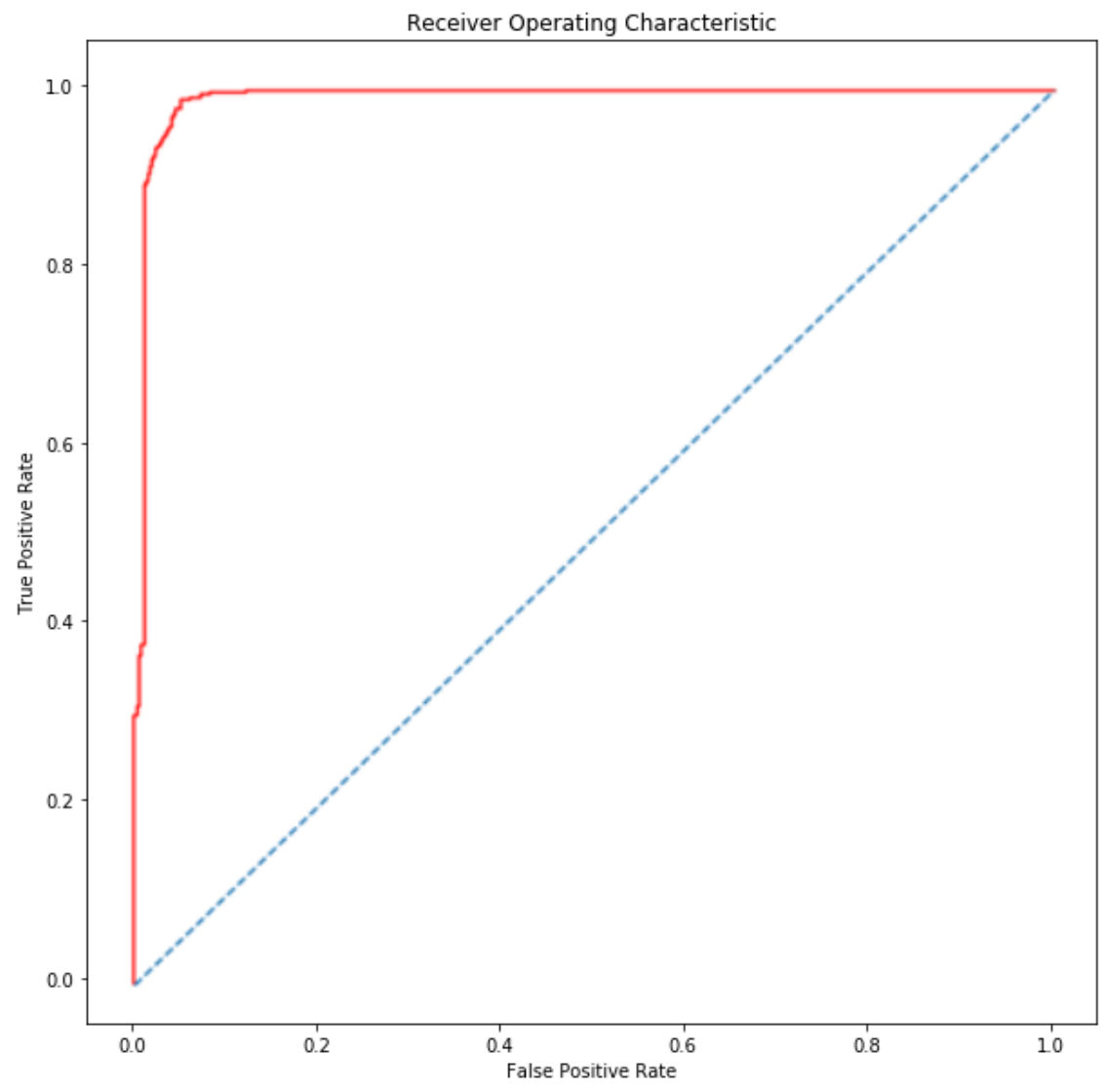

**(a)** The MiniRocket ROC curve of 10 subjects

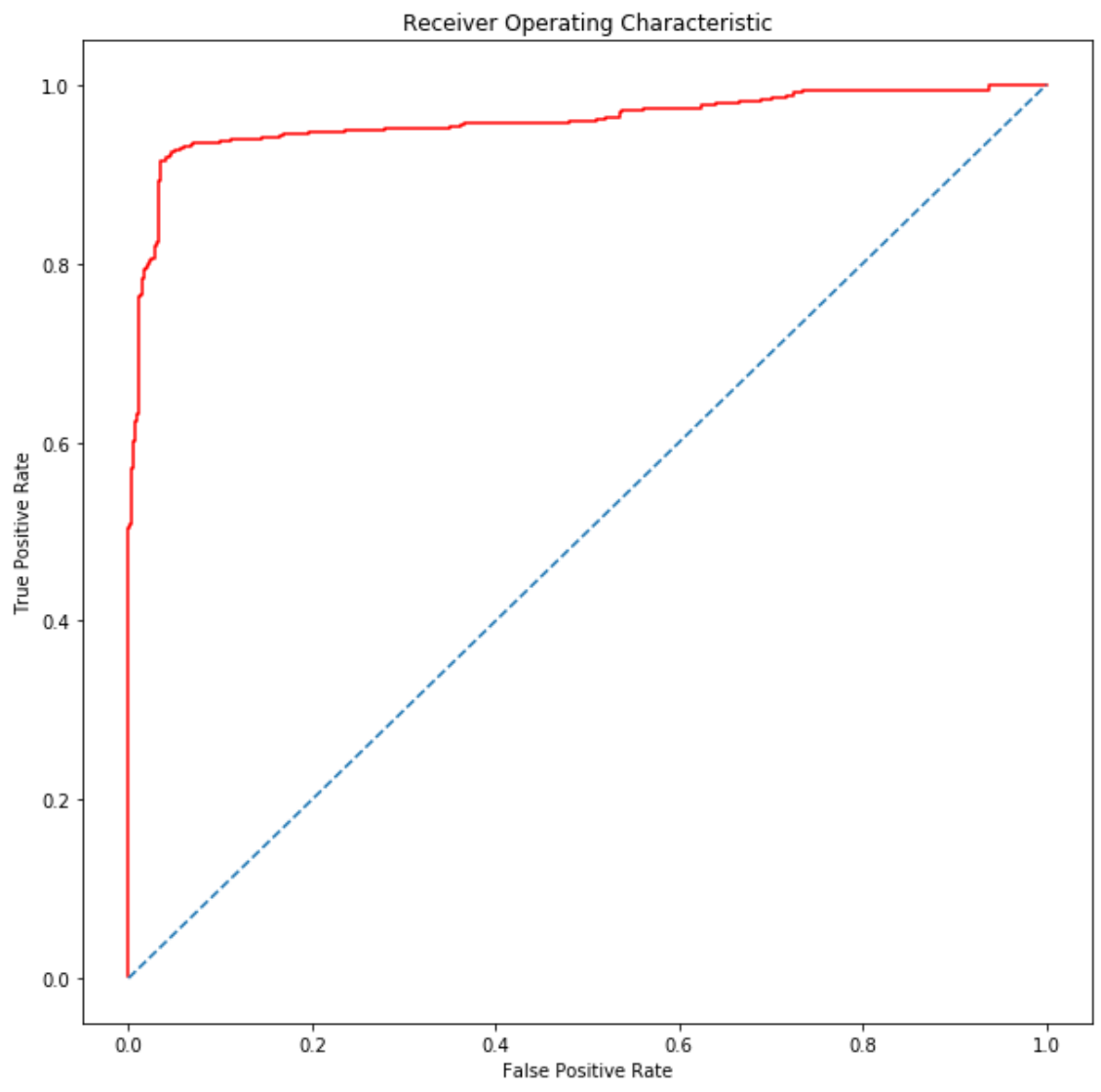

**(b)** The CNN-LSTM deep learning ROC curve of 10 subjects can be directly reflected in the ROC curve. When the curve is less than the oblique diagonal, the classifier is less accurate than the random classifier. Aside from that, the classifier is superior to the random classifier.

**Figure 8:** The ROC curves of all subjects in different models.

Fig 9 depicts the accuracy and loss during training and validation sessions of the proposed approaches on the PhysioNet dataset. The network is trained over 100 epochs using a batch size of 400. As can be seen, the proposed approaches achieve high accuracy on the public dataset for four different types of MI tasks, with scores of 99.29%, 97.12%, 96.12%, and 99.30%, respectively. The result reveals that the proposed algorithms can enhance the multi-classification MI tasks' accuracy.

The proposed models epoch versus accuracy and loss plots on the train and validation datasets. (a) and (b) display the accuracy and loss function curves for the MiniRocket model. (c) and (d) display the accuracy and loss function curves for the hybrid deep learning model.

## 5. Discussion

Numerous investigations have been carried out to enhance the classification accuracy of motor imagery tasks [93]. The temporal information contained in the EEG signals was used in these studies. Numerous frequency bands are present in EEG signals, and the biological significance of each of these bands varies. Specifically, $\mu$ and $\beta$ bands contain the most discriminating MI features. The most effective method for making use of these frequency characteristics and temporal data contained in EEG signals is through the use of time-frequency representation. Additionally, an EEG signal representation in two dimensions can speed up CNN learning due to the fact that these networks are the most effective networks to recognise spatial patterns in the images input. Nevertheless, adding more layers to a CNN requires a large dataset to train the massive parameters in auxiliary layers.

Table 2 shows our system's highest performance when comparing its overall performance (accuracy) to that of the best-performing earlier systems that used the PhysioNet dataset. These variations in the methods employed help to explain the causes of this higher performance.

Using the C3, C4, and CZ channels of EEG data, Alomari et al. [82] extracted features using a discrete wavelet transform (DWT) technique and used a support vector machine (SVM) to classify motor imagery movements. The overall accuracy for both the fist and foot MI tasks was 74.90%. The method developed in [76] employed the 64 channels raw input data and, using four classes, the average accuracy was 80.38%. Furthermore, it employed a 2D CNN methodology and a relatively straightforward neural network as opposed to the 1D approach used here and in [89].

**Table 2**
Comparison of results using the PhysioNet EEG dataset.

| Work | MI tasks | Accuracy | Methods |
|---|---|---|---|
| Alomari *et al.* [82] | 2 | 74.90% | SVM |
| Karácsony *et al.* [83] | 4 | 76.37% | CNN |
| Dose *et al.* [76] | 4 | 80.38% | CNN |
| Sita *et al.* [84] | 3 | 87.24% | LDA+RDA |
| Hou *et al.* [85] | 4 | 88.57% | GCNs-net |
| Hou *et al.* [86] | 4 | 94.50% | CNN |
| Hou *et al.* [87] | 4 | 94.64% | Bi-LSTM |
| Zhang *et al.* [88] | 5 | 95.53% | LSTM |
| Lun *et al.* [89] | 4 | 95.76% | CNN |
| Li *et al.* [90] | 5 | 96.41% | DSCNN+GRU |
| Li *et al.* [91] | 5 | 97.36% | CNN+GRU |
| Li *et al.* [92] | 5 | 97.71% | DSCNN+ELM |
| This work | 4 | 98.63% | MiniRocket |
| | | 98.06% | CNN-LSTM |

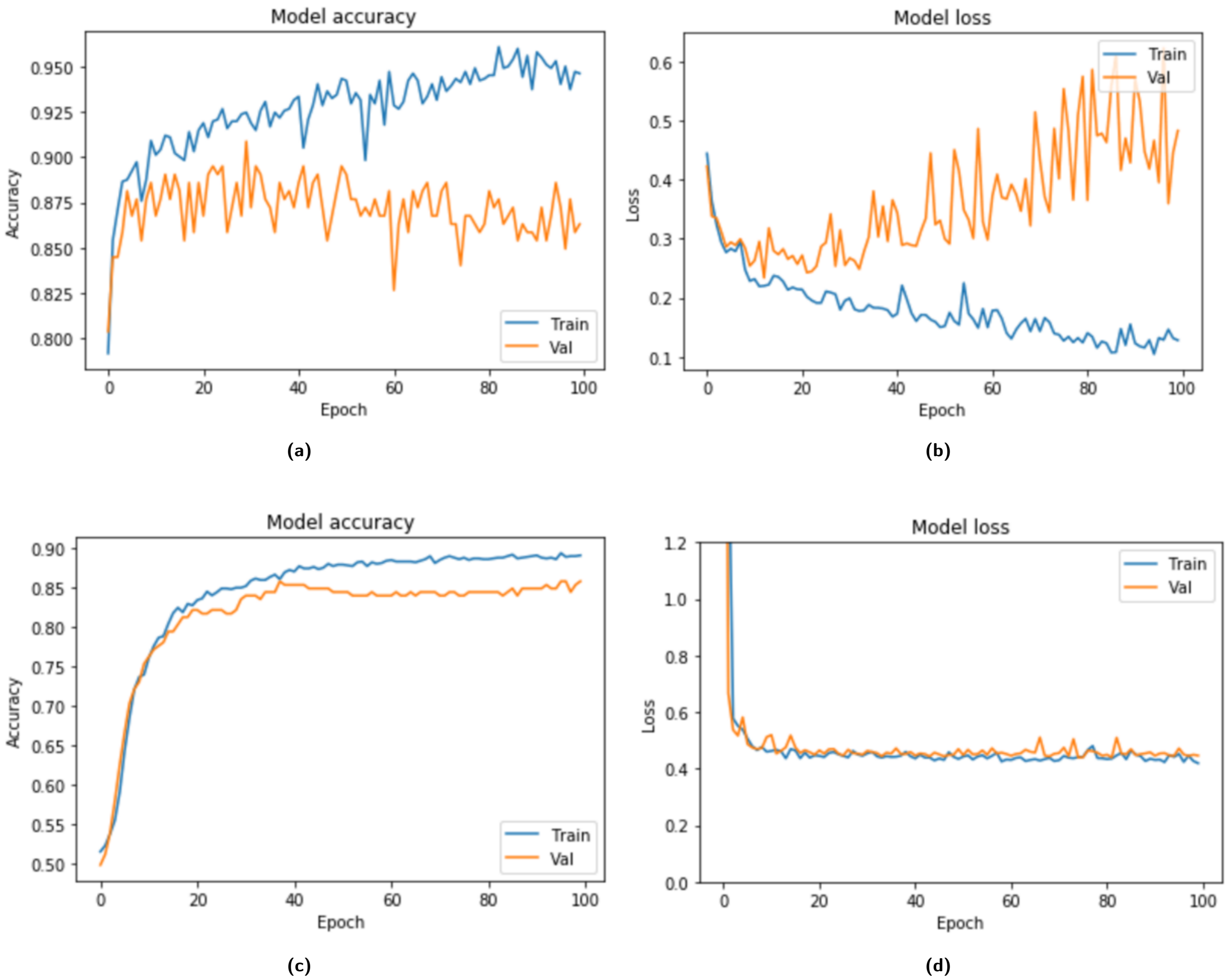

**Figure 9:** The proposed models epoch versus accuracy and loss plots on the train and validation datasets. (a) and (b) display the accuracy and loss function curves for the MiniRocket model. (c) and (d) display the accuracy and loss function curves for the hybrid deep learning model.

The method developed in [83] utilised various channel configurations and obtained the best performance using input data from all 64 channels. It achieved an average accuracy of 76.37% with four classes. Instead of using raw data as is the case here, it utilised pre-processed data, thus potentially losing important information. Furthermore, it employed a 2D CNN methodology and a relatively straightforward neural network as opposed to the 1D approach used here and in [89]. It did not employ spatial drop-out.

Gaussian (GS) weighting was utilised by Sita and Nair [84] for the features that were extracted for EEG signals through ICA. In order to classify the MI tasks, both linear discriminant analysis (LDA) and rapid discriminant analysis (RDA) were utilised as classifiers. For the MI tasks involving the right fist, the left fist, and both feet, the overall accuracy with 64 channels was 87.24%.

In order to fix the forward and inverse issues of EEG signals, Hou et al. [86] implemented the boundary element method (BEM) and weighted minimum norm estimation (WMNE) of EEG source imaging (ESI) technology. These helpful characteristics were extracted from the scout subsequence, which was generated in the motor cortex after 10 scouts were generated to select the ROI. Using data from 10 subjects performing MI tasks with their left fist, right fist, both fists, and both feet, CNN was able to classify the MI actions with a recognition rate of 94.50%.

Moreover, without any pre-processing or artificial feature extraction operations, Lun et al. [89] developed a deep CNN structure with separated temporal and spatial filters, which selected the raw EEG signals of the electrode pairs over the motor cortex region as hybrid samples. The results showed that the average global accuracy for classifying 10 and 100 subjects using either a left or right fist, or both fists and both feet, was 97.28% and 94.80%, respectively.

In the study by Zhang et al. [88], the authors constructed a seven-layer deep learning model that classified the initial EEG signals using a combination of LSTM and RNN. Simultaneously, hyperparameters were chosen using an orthogonal array experiment method. Lastly, the rate of intention recognition using EEG reached 95.53% for five types of MI tasks.

The attention mechanism strategy was used to investigate prospective features of EEG signals, and a bidirectional long-short term memory (Bi-LSTM) network was proposed by Hou et al. [87] for studying the recognition of MI tasks and obtained an adequate recognition rate of 94.64%.

Using the deep learning framework, Hou et al. [85] decoded EEG motion imagination signals as graph convolution, whose functional topology not only matched that of the electrodes but also bridged the gap left by the Euclidean structure of the electrodes, which was unable to precisely capture the connection between signals in its entirety, which not only matched the functional topology of electrodes but also bridged the gap between the signals' interactions, which the Euclidean structure of electrodes could not fully reflect. Moreover, the performance of decoding the original EEG signals was enhanced, as evidenced by an increased recognition rate of 88.57% for 4 MI tasks.

Li et al. [91] used the two-dimensional segment structure of spatial–temporal EEG data to build a CNN-Gate recurrent unit (GRU) network mixed depth neural network model. The EEG signals' spatial features were extracted using CNN, and their temporal features were extracted using GRU. Additionally, a comparison was made between recognition speed and channel choice. In the end, the classification accuracy for the MI tasks with eyes closed, left fist, right fist, both fists, and both feet achieved 97.36%. Furthermore, Li et al. [90] developed a hybrid network model structure that combines DSCNN and GRU to classify five types of EEG-related MI tasks. The intention recognition rate reached 97.76%.

Li et al. [92] developed a novel approach to enhance the MI classification by using deep separation convolutional neural networks (DSCNN) and extreme learning machines (ELM). The acquired EEG signals are initially preprocessed from a single dimension into a two-dimensional grid with spatial and temporal characteristics. The DSCNN is then used to separate temporal features from spatial features. Finally, the extracted temporal and spatial features are used by the ELM classifier to categorise five distinct MI tasks. The proposed framework achieved an outstanding intention recognition rate of 97.88% using the publicly available PhysioNet datasets.

## 6. Conclusions

This paper proposed a novel approach for the EEG classification of four motor imagery tasks based on MiniRocket and a CNN-LSTM hybrid neural network. High computational complexity is a limitation of the majority of state-of-the-art time series classification techniques. However, the use of MiniRocket helps in overcoming that barrier. The optimal number of CNN blocks for feature segmentation of the MI-EEG signal in a hybrid neural network, as well as evaluations, were conducted on the use of fully connected neural networks and pre-trained CNNs in MI-based systems, as well as on their combination with trainable LSTM.

In conclusion, the results of this study on the public dataset, PhysioNet, show that the proposed frameworks can distinguish between MI tasks within subjects with maximum computation efficiency and highest accuracy. Next steps will involve designing data collection experiments, adapting the approach to different experimental paradigms, and carrying out online experimentation.

## Footnotes

## CRediT authorship contribution statement

**Jamal Hwaidi:** Conceptualization; Methodology; Data Curation; Resources; Investigation; Visualization; Writing original draft; Writing, review, and editing. **Mohamed Chahine Ghanem:** Conceptualization; Methodology; Data Curation; Resources; Investigation; Visualization; Supervision ; Writing: review & editing.

### Acknowledgements

The open access (OA) fee for this paper was funded by the University of Liverpool.

### Declaration of Competing Interest

No competing financial interests or personal affiliations are associated with the authors that could have impacted the content of this paper.

### Data Availability

PhysioNet MI-EEG dataset [94] is publicly available at https://physionet.org/content/eegmmidb/1.0.0/. The data that support the findings of this study are available upon request from the corresponding author.